\documentclass{article}
\usepackage[preprint]{neurips_2025}
\usepackage[utf8]{inputenc} 
\usepackage[T1]{fontenc}    
\usepackage{hyperref}       
\usepackage{url}            
\usepackage{booktabs}       
\usepackage{amsfonts}       
\usepackage{nicefrac}       
\usepackage{microtype}      
\usepackage{xcolor}         
\usepackage{amsmath}
\usepackage{amssymb}
\usepackage{mathtools}
\usepackage{amsthm}
\usepackage[capitalize,noabbrev]{cleveref}
\usepackage{xspace}
\usepackage{hyperref}
\usepackage{enumitem}
\usepackage{url}
\usepackage{graphicx}
\usepackage{booktabs}
\usepackage[ruled]{algorithm2e}
\usepackage{algorithmic}
\usepackage{multirow}
\usepackage{colortbl}
\usepackage{wrapfig}
\usepackage{float}
\usepackage{makecell}
\usepackage{wrapfig}   
\usepackage{colortbl}
\usepackage{pifont}  
\newcommand{\cta}{\texttt{Dota}\xspace}
\newcommand{\CTA}{\texttt{DOTA}\xspace}

\theoremstyle{plain}
\newtheorem{theorem}{Theorem}[section]
\newtheorem{proposition}[theorem]{Proposition}

\theoremstyle{definition}

\newtheorem{assumption}[theorem]{Assumption}
\theoremstyle{remark}

\title{\CTA: DistributiOnal Test-time Adaptation of Vision-Language Models}

\author{Zongbo Han$^{1*}$, Jialong Yang$^2$, Guangyu Wang$^1$, Junfan Li$^3$,Qianli Xu$^4$, Mike Zheng Shou$^{5*}$,\\
\textbf{Changqing Zhang$^2$}\thanks{Corresponding author: zongbo@bupt.edu.cn, mike.zheng.shou@gmail.com, zhangchangqing@tju.edu.cn}
\\
State Key Laboratory of Networking and Switching Technology, Beijing University of \\ Posts and Telecommunications$^1$ \\
College of Intelligence and Computing, Tianjin University$^2$\\
School of Computer Science and Technology, Harbin Institute of Technology Shenzhen$^3$\\
Institute for Infocomm Research, A*STAR$^4$\\
Show Lab, National University of Singapore$^5$
}

\begin{document}

\maketitle
\begin{abstract}
\textcolor{black}{Vision-language foundation models (VLMs), such as CLIP, exhibit remarkable performance across a wide range of tasks. However, deploying these models can be unreliable when significant distribution gaps exist between training and test data, while fine-tuning for diverse scenarios is often costly. Cache-based test-time adapters offer an efficient alternative by storing representative test samples to guide subsequent classifications. Yet, these methods typically employ naive cache management with limited capacity, leading to severe catastrophic forgetting when samples are inevitably dropped during updates. In this paper, we propose \cta (DistributiOnal Test-time Adaptation), a simple yet effective method addressing this limitation. Crucially, instead of merely memorizing individual test samples, \cta continuously estimates the underlying distribution of the test data stream. Test-time posterior probabilities are then computed using these dynamically estimated distributions via Bayes' theorem for adaptation. This distribution-centric approach enables the model to continually learn and adapt to the deployment environment. Extensive experiments validate that \cta significantly mitigates forgetting and achieves state-of-the-art performance compared to existing methods.}
\end{abstract}

\section{Introduction}
\textcolor{black}{Recent advances in vision-language foundation models have shown remarkable vision understanding capabilities across a broad range of tasks by training on web-scale image-text pairs \citep{radford2021learning,DBLP:conf/icml/LavoieKIAWCB24,zhai2023sigmoid}. Taking CLIP as an example, it can conduct zero-shot classification without the need for additional training data using predefined prompts \citep{radford2021learning}. However, CLIP may still face challenges when handling various specific applications during test time, especially when there is a significant distribution gap between the training and test data \citep{shu2022test,karmanov2024efficient,feng2023diverse}. To adapt a foundational model to diverse deployment environments and personalized application requirements, fine-tuning is often necessary, which can be resource-intensive in terms of time, computational effort, and training data requirements.}

\textcolor{black}{Test-time adaptation (TTA) methods provide an efficient solution for addressing distributional shifts between training and testing domains \citep{boudiaf2022parameter, chen2022contrastive, DBLP:conf/iclr/WangSLOD21}. TTA allows for the dynamic adjustment of a pre-trained model during the inference phase by leveraging incoming test data to refine the model's parameters. This adaptation process optimizes the model’s performance for the specific test distribution, eliminating the need for resource-intensive retraining. As such, TTA aligns seamlessly with real-world applications, where models must rapidly adapt to diverse and changing environments. There are two primary lines to achieve TTA on VLMs. Early works advocate learning prompts during test time with the test data \citep{shu2022test,feng2023diverse}. However, these methods require significant computational resources to optimize the learnable prompts via backpropagation and gradient descent, rendering them unsuitable for applications demanding fast test-time inference. Therefore, more efficient methods, cache-based methods, have been proposed recently \citep{karmanov2024efficient, zhang2024boostadapter}. Typically, to avoid the need for training with backpropagation, Training-free Dynamic Adapter (TDA) maintains a lightweight cache during testing to store representative test samples, which helps guide the classification of subsequent samples.}


\begin{wrapfigure}[25]{r}{0.45\textwidth}
    \centering
\includegraphics[width=\linewidth,trim=10 10 10 10]{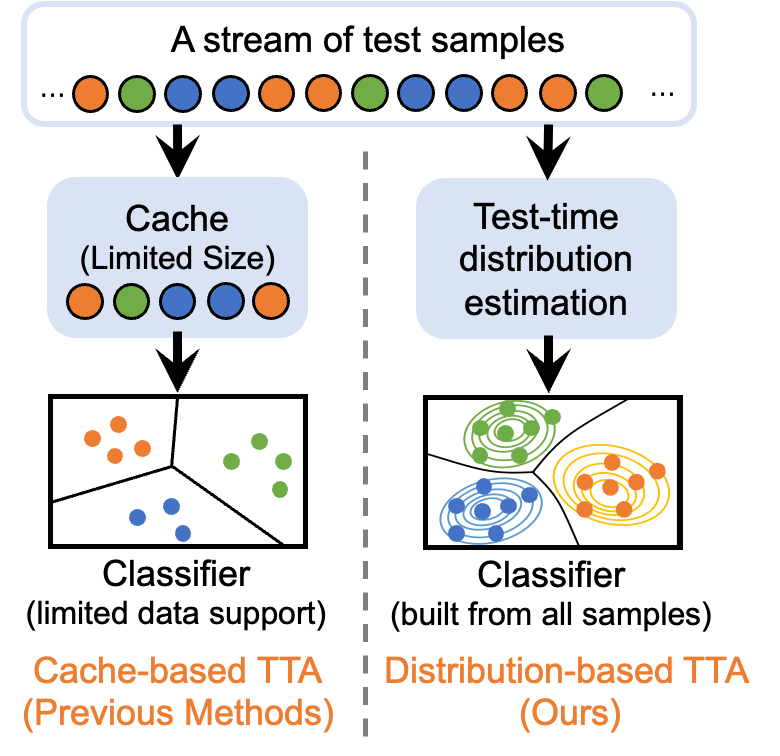} 
\vspace{-1em}
    \caption{Cache-based TTA methods store individual test samples within a limited cache, which often leads to underutilization of the available test data. In contrast, the proposed method continuously estimates the underlying distribution of the test data, enabling the full exploitation of all available test samples.}
    \label{fig:motivation}
\end{wrapfigure}

\textcolor{black}{While cache-based classifiers like TDA offer significant efficiency, they are fundamentally constrained by their finite cache capacity. These methods typically store a limited set of `typical' test samples, updating the cache by replacing older entries with newer, high-confidence ones. However, this reliance on instance-level memorization and forced sample discarding inevitably leads to catastrophic forgetting. As the model adapts to new data patterns, it loses information about previously seen variations, hindering the formation of a stable and comprehensive understanding of the evolving test distribution. Consequently, classifiers depending solely on cached samples can be suboptimal, and ultimately performance-limited by the cache size \cite{zhang2024boostadapter}. As illustrated in Fig.~\ref{fig:motivation}, to overcome these limitations of instance-level caching, we introduce DistributiOnal Test-time Adaptation (\cta). This represents a paradigm shift in test-time adaptation. Instead of passively memorizing discrete samples in the cache with limited size, the core motivation of \cta lies in actively and continuously estimating the underlying statistical distribution of the incoming test data, thereby enabling the full utilization of all available test samples.}

\textcolor{black}{Specifically, \cta continually estimates the distribution of test samples to adapt the test environment. Under the mild assumption that the embedding distribution of each class follows a Gaussian distribution \citep{hastie1996discriminant}, we propose an efficient method to continually estimate the distribution of different classes. Once the distributions of different classes are estimated, we can easily calculate the posterior probabilities of subsequent test samples based on Bayes' theorem and obtain a test-time classifier for test-time adaptation. Similar to cache-based methods, this process does not require gradient backpropagation, avoiding the complex computational overhead during testing, leading to more than 20 times faster inference speed. Moreover, unlike cache-based methods memorizing only representative test samples, \cta can continually adapt to the test environment by estimating the distribution of test samples. 
The contributions of this paper are:
\begin{itemize}
    \item We propose a novel continual test-time  learning framework which improve the performance of pretrained foundation models in downstream tasks. 
    \item Within this framework, we propose a simple yet effective method to enhance the foundation model by efficiently estimating the distribution of different categories during test time. 
    \item Extensive experiments on diverse datasets validate the effectiveness of the proposed method, demonstrating a significant improvement. The code will be released.
\end{itemize}}
\section{Related work}
\textcolor{black}{\textbf{Test-time adaptation (TTA)} for classical classification neural networks focuses on addressing the distribution shift between training and test data by learning from the test data. Early efforts to improve TTA performance primarily involve adjusting batch normalization layers and designing unsupervised objective functions \citep{nado2020evaluating,wang2020tent,khurana2021sita,lim2023ttn}. For example, TENT \citep{wang2020tent} optimizes the affine parameters in batch normalization layers by minimizing the entropy of the prediction probability. MEMO \citep{zhang2022memo} applies variant augmentation methods to a single test sample and optimizes model parameters by minimizing the entropy of the prediction probability. To enable practical test-time adaptation in dynamic, time-correlated test data streams, such as autonomous driving, RoTTA introduces novel robust batch normalization, a memory bank for balanced sampling, and a time-aware reweighting strategy\citep{yuan2023robust}. A recent advancement in test-time adaptation with distribution shift, which introduces the concept of universal TTA to address domain non-stationarity and temporal correlation, ensuring robust model performance across diverse scenarios \citep{marsden2024universal}. The most relevant of these traditional TTA methods to us is T3A \citep{iwasawa2021test}, which achieves test-time adaptation by adjusting the trained linear classifier using prototypes. Compared to T3A, which naively stores typical test samples, we achieve continuous adaptation by estimating the distribution of test samples.}

\textcolor{black}{\textbf{TTA for VLMs}. {To enhance the performance of VLMs during testing, TPT \citep{shu2022test} introduces adaptive text prompts and optimizes the prompts through entropy minimization. Building on this, DiffTPT \citep{feng2023diverse} leverages pre-trained stable diffusion models to generate diverse augmented data for use in test-time prompt tuning. However, TPT and DiffTPT rely heavily on gradient backpropagation to optimize the prompts, making them computationally expensive and resource-intensive. TDA \citep{karmanov2024efficient} proposes a lightweight test-time adaption method by storing representative test samples. Building upon TDA, Boostadapter \citep{zhangboostadapter} enhances the cache sample selection strategy through Regional Bootstrapping. Compared to TDA and Boostadapter, which naively stores typical test samples, we achieve continuous adaptation by estimating the distribution of test samples, leading to a more adaptive solution.}}

\textbf{Distribution estimation for recognition.} Distribution estimation plays a crucial role in adapting recognition models by leveraging data's statistical properties for dynamic updates. This approach is particularly effective when encountering new classes or shifting data distributions \citep{hastie1996discriminant}. For example, \cite{bendale2015towards} developed an open-world recognition system that continuously learns new object categories by evolving the nearest class mean algorithm into a nearest non-outlier variant. Similarly, Prototypical Networks \citep{snell2017prototypical} utilize distribution estimation by defining class prototypes as the mean of embedded support examples; classification then relies on metric distances, yielding strong performance in few-shot and zero-shot settings. Further addressing dynamic data, \cite{de2021continual} proposed a system for continual prototype evolution, facilitating online learning from non-stationary streams via efficient memory management and a novel objective function. Building upon these principles, particularly from the continual learning literature, this paper introduces \cta to enhance the test-time performance of vision-language models.


\noindent \textbf{Vision-language models} have demonstrated strong vision understanding capabilities benefiting from training on large-scale datasets \citep{radford2021learning,zhai2023sigmoid,DBLP:conf/icml/LavoieKIAWCB24}. Among them, CLIP \citep{radford2021learning} is the most representative method by maximizing the similarity between image and their corresponding text embeddings. To further enhance performance of CLIP on downstream tasks, prompt learning-based methods have been proposed by optimizing the prompts of the text encoder \citep{zhou2022cocoop,zhou2022coop,bai2024id,khattak2023maple,bai2024id}. Moreover, to reduce the computational cost associated with gradient calculations in prompt learning, efficient CLIP adaptation methods have been introduced \citep{gao2024clip,zhang2022tip,DBLP:conf/iclr/WangLS0WT24,li2024graphadapter,yu2023task}. These methods enable downstream task adaptation using only a small number of training samples in the embedding space. Orthogonal to above methods, this paper focuses on continuously adapting to environments during testing by leveraging test samples.

\section{Method}
\subsection{Zero-Shot Classification with Prompt}

\textbf{Zero-shot classification.} \textcolor{black}{During the pre-training stage, CLIP\footnote{\cta is also applicable to other similar models.} trains its image and text encoders using large-scale image-text pairs. This is achieved by maximizing the cosine similarity between the image and text embeddings through contrastive loss. Unlike traditional classifiers trained on closed-set labels, CLIP leverages open-set semantic information in the image-text pairs to learn a broader range of visual concepts. Consequently, during the test stage, CLIP can perform zero-shot classification without additional training. Specifically, given a test sample \( \boldsymbol{x} \) for \( K \)-class classification, where \( \boldsymbol{x} \) represents the image embedding obtained from the image encoder, the corresponding zero-shot prediction probability $P^{\texttt{zs}}_{k}$ for class $k$ is calculated as:
\begin{equation}
\label{eq:zeroshot}
P^{\texttt{zs}}_{k}(y=k|\boldsymbol{x})=\frac{\exp(\cos(\boldsymbol{x},\boldsymbol{w}_k)/\tau)}{\sum_{k=1}^K\exp(\cos(\boldsymbol{x},\boldsymbol{w}_k)/\tau)},    
\end{equation}
where $\texttt{zs}$ refers to \textbf{\texttt{z}}ero-\textbf{\texttt{s}}hot. \( \boldsymbol{w}_k \) is the classification weight for class \( k \), obtained by encoding the corresponding prompt, e.g., “a photo of \{class\}”, with the class token replaced by the specific category name. \( \tau \) is the learned temperature parameter in CLIP, and $\cos(\cdot, \cdot)$ denotes the cosine similarity. The above classification process can be understood as comparing the obtained image embedding with the text prompt and selecting the most similar category as the final decision.}

\begin{figure*}[!tbp]
\centering
\includegraphics[width=1\textwidth]{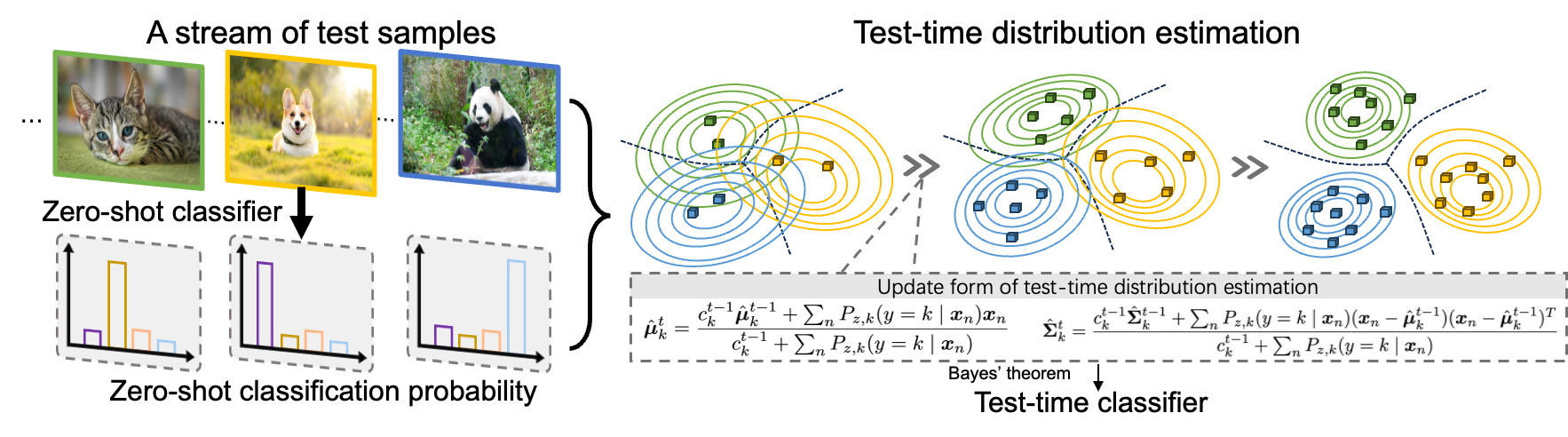} 
\caption{Pipeline of \cta. During test time, a stream of test samples is evaluated with original zero-shot classifier, and we estimate the distributions for the test samples during testing, enabling the model to continually learn from the test samples and the zero-shot classification probabilities. \textit{As the number of test samples increases, the estimated test sample data distribution will become more accurate.} Finally the test-time classifier can be obtained with the estimated distributions according to Bayes' theorem for test-time adaptation.}
\label{fig1:framework}
\end{figure*}
\subsection{Distributional Test-time Adaptation}
\label{sec:ctta}
\textcolor{black}{\textbf{Key motivation.} When CLIP is deployed across different environments, its performance often deteriorates due to changes in the data distribution, especially when the test data significantly deviates from the training data. TTA can effectively enable the foundational model to adapt quickly to new environments during the test phase. Current state-of-the-art methods typically maintain a cache of representative samples from various classes, which guide the classification of subsequent test samples \cite{karmanov2024efficient, zhang2024boostadapter}. However, due to the limited cache size, cache-based TTA methods face several challenges:
\begin{itemize}
    \item Limited information storage: These methods store embeddings for only a limited number of samples, which restricts the breadth and depth of information captured. As a result, they may fail to adequately represent the full complexity and nuances of the test data distribution.
    \item Lack of deep association learning: These methods primarily rely on cached samples for similarity matching or guidance, rather than learning or updating the model's understanding of the deeper, intrinsic relationships between sample features and their corresponding semantic labels in the new test environment.
    \item Significant risk of test-time forgetting: Due to the above limitations, when the cache must be updated (e.g., by replacing older samples), the model struggles to retain the adaptive knowledge gained previously. This leads to a high risk of catastrophic forgetting, compromising both performance and stability as new data arrives.
\end{itemize}
To address these challenges, as shown in Fig.~\ref{fig:motivation}, we propose distributional test-time adaptation (\cta), which continuously estimates the evolving test sample distribution during testing. By leveraging Bayes' theorem to infer the posterior distribution of different classes, \cta allows for dynamic adaptation based on an accurate understanding of the current data distribution. This distributional estimation ensures richer, more reliable information for classification, overcoming the limitations of cache-based methods and reducing the risk of forgetting, thus maintaining stable performance.}

\textcolor{black}{\textbf{Classification with classical Gaussian discriminant analysis.}{Formally, inspired by classical Gaussian discriminant analysis \citep{hastie1996discriminant}, we assume that the embedding distribution of each class $k$ follows a Gaussian distribution, i.e., $P(\boldsymbol{x}\!\!\mid\!\!y\!\!=\!\!k)=\mathcal{N}(\boldsymbol{\mu}_k, \boldsymbol{\Sigma}_k)$, where $\boldsymbol{\mu}_k$ and $\boldsymbol{\Sigma}_k$ are the mean vector and covariance matrix of class $k$, respectively.
Using Bayes’ theorem, the posterior probability $P(y\!=\!k|\boldsymbol{x})$ of class $k$ can be given by 
\[P(y\!=\!k|\boldsymbol{x}) = \frac{P(\boldsymbol{x}|y\!=\!k) P(y\!=\!k)}{P(\boldsymbol{x})},\]
where $P(\boldsymbol{x})=\sum_{k=1}^K P(\boldsymbol{x}|y\!=\!k)P(y\!=\!k)$ and $P(y\!=\!k)$ is the prior probability. 
In practice, we set $P(y\!=\!k)$ to $1/K$ for simplicity. 
Then $P(y\!=\!k |\boldsymbol{x})$ can be obtained with
\begin{equation}
\label{eq:testprob_qda}
P(y\!=k\mid\!\boldsymbol{x})=\frac{\exp(f_k(\boldsymbol{x}))}{\sum_{k=1}^K \exp(f_k(\boldsymbol{x}))},    
\end{equation}
where $f_k(\boldsymbol{x})=-\tfrac{1}{2}(\boldsymbol{x}-\boldsymbol{\mu}_k)^T\boldsymbol{\Sigma}_k^{-1}(\boldsymbol{x}-\boldsymbol{\mu}_k) - \tfrac{1}{2}\log |\boldsymbol{\Sigma}_k|$.} The discriminant function $f_k(\boldsymbol{x})$ measures how well a sample $\boldsymbol{x}$ fits the distribution of class $k$. The detail can be found in the Appendix~\ref{sec:fkxdetail}.
For the \( k \)-th class with \( N_k \) training samples, the mean and covariance matrix of different classes can be estimated as follows:
\begin{equation}
\label{ea:estimation_gda}    
\hat{\boldsymbol{\mu}}_k = \frac{\sum_{n=1}^{N_k} \boldsymbol{x}_n}{N_k}, 
\hat{\boldsymbol{\Sigma}}_k = \frac{\sum_{n=1}^{N_k} (\boldsymbol{x}_n - \hat{\boldsymbol{\mu}}_k)(\boldsymbol{x}_n - \hat{\boldsymbol{\mu}}_k)^T}{N_k},
\end{equation}
where \( \hat{\boldsymbol{\mu}}_k \) and \( \hat{\boldsymbol{\Sigma}}_k \) represent the estimated mean and covariance matrix of the \( k \)-th class, respectively. $\boldsymbol{x}_n$ denotes the input embeddings. However, there are two problems with the estimation of the mean and covariance in Eq.~\ref{ea:estimation_gda} that make it unsuitable for test-time distribution estimation. First, during testing, class labels are not available to compute the mean and covariance for class \(k\). Second, during testing, it is impossible to access all test samples at once for distribution estimation, as the test samples are provided in the form of a data stream. We will address these two issues one by one below.}

\textcolor{black}{\textbf{Parameter estimation with zero-shot predictive probability.} When conducting test-time distribution estimation, one main challenge is that we cannot access to the ground-truth labels for the $N$ test samples. Therefore, we try to use the zero-shot predictive probability to estimate the distribution \citep{hastie1996discriminant}. To this end we introduce the Prop.~\ref{tta:lemma1}. The proof can be found in the appendix~\ref{proof:tta:lemma1}.}
\textcolor{black}{\begin{proposition}[Parameter estimation with zero-shot predictive probability using the Expectation-Maximization (EM) algorithm]
\label{tta:lemma1}
Let \(\{P^{\texttt{zs}}_{k}(y\!=k\mid \boldsymbol{x}_n)\}_{k=1}^K\) be the zero-shot prediction probabilities for \(n\)-th test sample. The means \(\{\hat{\boldsymbol{\mu}}_k\}_{k=1}^K\) and covariances \(\{\hat{\boldsymbol{\Sigma}}_k\}_{k=1}^K\) of the distribution can be estimated as a single iteration of the EM algorithm, where the expectation step computes the zero-shot prediction and the maximization step updates the parameters by maximizing the likelihood. The estimates are:
\begin{equation}
\begin{aligned}
\label{eq:estimation1}
\hat{\boldsymbol{\mu}}_k = \frac{\sum_{n=1}^N P^{\texttt{zs}}_{k}(y\!=k\mid \boldsymbol{x}_n) \boldsymbol{x}_n}{\sum_{n=1}^N P^{\texttt{zs}}_{k}(y\!=k\mid \boldsymbol{x}_n)}, 
\hat{\boldsymbol{\Sigma}}_k = \frac{\sum_{n=1}^N P^{\texttt{zs}}_{k}(y\!=k\mid \boldsymbol{x}_n) (\boldsymbol{x}_n - \hat{\boldsymbol{\mu}}_k)(\boldsymbol{x}_n - \hat{\boldsymbol{\mu}}_k)^T}{\sum_{n=1}^N P^{\texttt{zs}}_{k}(y\!=k\mid \boldsymbol{x}_n)}.
\end{aligned}
\end{equation}
\end{proposition}}
\textcolor{black}{Prop.~\ref{tta:lemma1} shows that the estimation process with zero-shot predictive probability. The estimation can also be intuitively understood as reweighting, where the zero-shot probabilities are used as weights to adjust the contributions of different samples, thereby mitigating the impact of the potential inaccuracies in the zero-shot prediction.}


\textcolor{black}{\textbf{Online test-time distribution estimation.} When estimating data distribution at test time, one another challenge is that we evaluate the test samples sequentially in a streaming manner instead of accessing all samples simultaneously. This necessitates a strategy to appropriately adjust the estimation method in Eq.~\ref{eq:estimation1} through effective initialization, and then allowing the parameters to be updated quickly as new test samples arrive. To achieve this goal, \cta maintains the distribution information of different classes (i.e., mean and covariance matrix) during testing, and updates its distribution information based on its representation information after obtaining new samples.
\textbf{Initialization of \(\{\hat{\boldsymbol{\mu}}_k, \hat{\boldsymbol{\Sigma}}_k\}_{k=1}^K\)}. We can initialize the estimated mean of different classes in the following way:
\begin{equation}
\label{eq:init}
\hat{\boldsymbol{\mu}}_k^0 = \omega \mathbf{1} \quad \text{and} \quad \hat{\boldsymbol{\Sigma}}_k^0 = \sigma^2 \boldsymbol{I},
\end{equation}
where \(\omega\) \text{and} \(\sigma^2\) is a hyperparameter that determines the initial mean and variance, \(\boldsymbol{I}\) is the identity matrix. 
\textbf{Update of \(\{\hat{\boldsymbol{\mu}}_k, \hat{\boldsymbol{\Sigma}}_k\}_{k=1}^K\)}. We employ the update form described in \citep{dasgupta2007line}, which is capable of estimating Gaussian distribution parameters in an online setting. Theoretically, for any sequence, the average regret of the update form converges to zero in the limit. Specifically,  given a batch of test samples at step \(t\), the updated \(\hat{\boldsymbol{\mu}}_k^{t}, \hat{\boldsymbol{\Sigma}}_k^{t}\) can be computed based on the \(\hat{\boldsymbol{\mu}}_k^{t-1}, \hat{\boldsymbol{\Sigma}}_k^{t-1}\) as follows:
\begin{equation}
\begin{aligned}
\label{eq:update}
&\hat{\boldsymbol{\mu}}_k^{t} = \frac{c^{t-1}_k \hat{\boldsymbol{\mu}}_k^{t-1}+\sum P^{\texttt{zs}}_{k}(y=k\mid \boldsymbol{x}_n)\boldsymbol{x}_n}{c^{t-1}_k+\sum P^{\texttt{zs}}_{k}(y=k\mid\boldsymbol{x}_n)}
, \hat{\boldsymbol{\Sigma}}_k^{t} = \frac{c^{t-1}_k \hat{\boldsymbol{\Sigma}}_k^{t-1} + \sum P^{\texttt{zs}}_{k}(y=k\mid \boldsymbol{x}_n) \boldsymbol{S}_k^{t-1}}{c^{t-1}_k+\sum P^{\texttt{zs}}_{k}(y=k\mid \boldsymbol{x}_n)}\\
\end{aligned}
\end{equation}
where $\boldsymbol{S}_k^{t-1} = (\boldsymbol{x}_n - \hat{\boldsymbol{\mu}}_k^{t-1})(\boldsymbol{x}_n - \hat{\boldsymbol{\mu}}_k^{t-1})^T$, \(c_k^{t-1}\) represents the effective sample size, defined by the cumulative confidences of the observed samples of class \(k\) at step \(t-1\). Specifically, we set \(c_k^{0}=0\) and \(c_k^{t}\) is updated as \(c_k^{t}=c_k^{t-1} + \sum P^{\texttt{zs}}_{k}(y\!\!=\!\!k| \boldsymbol{x}_n)\). When we obtain the estimated \(\hat{\boldsymbol{\mu}}, \hat{\boldsymbol{\Sigma}}\), we can use Eq.~\ref{eq:testprob_qda} to calculate the test-time adapted posterior probability.}

\textcolor{black}{In practice, Eq.~\ref{eq:update} is a generalized vector update version that works effectively with different test batch sizes. For consistency with comparison methods, we set the batch size to 1 in our experiments. To reduce computational complexity when inverting the covariance matrix \(\hat{\boldsymbol{\Sigma}}_k\), similar to the approach in \citep{anderson1958introduction, friedman1989regularized}, we approximate the covariance by averaging across all classes, reducing the number of matrix inversions from \(K\) to \(1\), thereby improving efficiency and reducing the number of parameter estimates. Additionally, we apply shrinkage regularization to the precision matrix to enhance the stability of the inversion process as follows: \(\hat{\boldsymbol{\Lambda}} = [(1 - \epsilon) \hat{\boldsymbol{\Sigma}} + \epsilon \boldsymbol{I}]^{-1}\), where \(\epsilon=10^{-4}\) is the shrinkage parameter. The term \(\epsilon \boldsymbol{I}\) ensures that the eigenvalues of the covariance matrix are well-conditioned, maintaining the desired properties such as positive definiteness and rank stability.}

\subsection{Adaptive fusion of zero-shot and test-time classifier}
\textcolor{black}{As the number of test samples increases, the reliability of the estimated test sample distribution improves \citep{dasgupta2007line}. However, when the number of test samples is insufficient, the estimated distribution may be unreliable. To address this, we introduce a dynamic zero-shot classification and test-time result fusion approach, allowing the model to rely more on zero-shot classification during stages where the sample size for distribution estimation is insufficient. Formally, the final fusion probability is defined as follows:
\begin{equation}
\label{eq:fusion}
P_{k}(y=k|x)=\frac{\exp(\cos(\boldsymbol{x},\boldsymbol{w}_k)/\tau+\lambda f_k(\boldsymbol{x}))}{\sum_{k=1}^K[\exp(\cos(\boldsymbol{x},\boldsymbol{w}_k)/\tau+\lambda f_k(\boldsymbol{x}))]},
\end{equation}
where \(\lambda = \min(\rho c, \eta)\). Here, \(c\) represents the number of test samples, and \(\rho\) and \(\eta\) are hyperparameters that control the weight of the test-time classifier logits. The value of \(\lambda\) increases with the number of test samples when this number is insufficient, gradually approaching the maximum value \(\eta\). This approach encourages the model to rely on the zero-shot classifier results when the test samples are insufficient to estimate the distribution, mitigating the potential negative impact of the test-time classifier. The whole pseudo code is shown in Alg.~\ref{algo:cta}. }

\begin{algorithm}[!h]
\begin{algorithmic}
\caption{\textcolor{black}{The pseudocode of  \cta.}\label{algo:cta}}
        \STATE {\bfseries Input:}
            The embedding of $N$ test samples $\{\boldsymbol{x}_n\}_{n=1}^N$ in an streaming way, zero-shot classification weights $[\boldsymbol{w}_1,\cdots,\boldsymbol{w}_K]$;
        
        Initializing the distribution of different class;\\
        \FOR{each test sample $\boldsymbol{x}_i$}
        \STATE Obtain the zero-shot  probability with Eq.~\ref{eq:zeroshot};
        \STATE Update the distribution of different class with Eq.~\ref{eq:update};
        \STATE Obtain the test-time classification probability with Eq.~\ref{eq:testprob_qda};
        \STATE Obtain the final classification result with Eq.~\ref{eq:fusion}.
        \ENDFOR
\end{algorithmic}
\end{algorithm}

\section{Experiments}
\begin{table*}[!t]
  \caption{Top-1 accuracy (\%) under the cross-domain generalization scenario. ``PE'' represents the use of prompt enhancement\cite{pratt2023does}.}
  \centering
  \resizebox{0.97\linewidth}{!}{
    \begin{tabular}{lcccccccccc>{\columncolor{gray!20}}c}
      \toprule
{Method} & 
\rotatebox{90}{{Aircraft}} & 
\rotatebox{90}{{Caltech101}} & 
\rotatebox{90}{{Cars}} & 
\rotatebox{90}{{DTD}} & 
\rotatebox{90}{{EuroSAT}} & 
\rotatebox{90}{{Flower102}} & 
\rotatebox{90}{{Food101}} & 
\rotatebox{90}{{Pets}} & 
\rotatebox{90}{{SUN397}} & 
\rotatebox{90}{{UCF101}} & 
\rotatebox{90}{{Average}} \\
      \midrule
      Zero-Shot & 23.22  & 93.55  & 66.11  & 45.04  & 50.42  & 66.99  & 82.86  & 86.92  & 65.63  & 65.16  & 64.59  \\
      \midrule
    TPT\cite{shu2022test} & 24.78 & {94.16} & 66.87 & {47.75} & 42.44 & 68.98 & 84.67 & 87.79 & 65.50 & 68.04 & 65.10 \\
    DiffTPT\cite{feng2023diverse} & {25.60} & 92.49 & 67.01 & 47.00 & 43.13 & 70.10 & {87.23} & 88.22 & 65.74 & 62.67 &65.47 \\
    TDA\cite{karmanov2024efficient} & 23.91 & {94.24} & {67.28} & 47.40 & {58.00} & {71.42} & 86.14 & {88.63} & {67.62} & {70.66} & {67.53} \\
    BoostAdapter\cite{zhangboostadapter} & 27.45 & 94.77 & 69.30 & 45.69 & 61.22 & 71.66 & 87.17 & 89.51 & 68.09 & 71.93 & 68.68 \\
    HisTPT\cite{zhang2024historical} & 26.90 & 94.50 & 69.20 & 48.90 & 49.70 & 71.20 & 89.30 & 89.10 & 67.20 & 70.10 & 67.60 \\
    ZERO\cite{farina2024frustratingly} & 25.21 & 93.66 & 68.04 & 46.12 & 34.33 & 67.68 & 86.53 & 87.75 & 65.03 & 67.77 & 64.21 \\
     \cta &  {26.25}  &  {94.16} &  {69.56} &   {47.64} &  {62.78} &  {75.23} &  {87.08}  &  {92.01} &  {69.80} &  {72.54} &  {69.71} \\
     \midrule
     DMN w/PE \cite{zhang2024dual} & 30.03& 95.38& 67.96& 55.85& 59.43& 74.49& 85.08& 92.04& 70.18           & 72.51& 70.30 \\
\cta w/ PE & 29.82& 94.85& 69.06& 55.97         & 58.35            & 77.06           & 87.07            & 92.40         & 70.97           & 74.86         & 71.04           \\ 
    \bottomrule
    \end{tabular}
  } 
  \label{tab:fine-grained}
\end{table*}

\textbf{Benchmarks.} Consistent with prior works \citep{shu2022test,feng2023diverse,karmanov2024efficient}, we conduct our main experiments on cross-domain generalization and natural distribution shifts (NDS) scenarios. In the cross-domain generalization scenario, we evaluate the performance of the model across 10 diverse image classification datasets, each representing a distinct domain with different classes: Aircraft \citep{maji2013fine}, Caltech101 \citep{fei2004learning}, Cars \citep{krause20133d}, DTD \citep{cimpoi2014describing}, EuroSAT \citep{helber2019eurosat}, Flower102 \citep{nilsback2008automated}, Food101 \citep{bossard2014food}, Pets \citep{parkhi2012cats}, SUN397 \citep{xiao2010sun}, and UCF101 \citep{soomro2012dataset}. This benchmark provides a comprehensive evaluation of the adaptability of the model during test time across various class spaces. For the natural distribution shifts scenario, we utilize multiple datasets including ImageNet \citep{deng2009imagenet}, ImageNet-A \citep{hendrycks2021natural}, ImageNet-R \citep{hendrycks2021many}, ImageNet-S \citep{wang2019learning} and ImageNet-V2 \citep{recht2019imagenet}, which serve as measures of the robustness of our approach. We also evaluate \cta using medical image datasets and a pathology-pretrained foundation model \cite{huang2023visual}, where the labels were converted into descriptive sentences for evaluation (e.g., transforming ``tumor'' into ``H\&E image of a tumor''). The evaluation includes three datasets: the Kather Colon dataset \cite{kather2019predicting}, comprising nine different tissue types; the PanNuke dataset \cite{gamper2019pannuke}, focusing on benign versus malignant classifications; and the WSSS4LUAD dataset \cite{han2022wsss4luad}, which distinguishes between tumor and normal samples.

\textbf{Comparison Method.} We compare the proposed method with the following approaches: (1) CLIP's zero-shot method, which utilizes an ensemble of 80 prompts \citep{radford2021learning}; (2) Prompt-based training methods, including TPT \citep{shu2022test}, DiffTPT \citep{feng2023diverse}, and Historical Prompt Tuning \citep{zhang2024historical}. These methods focus on optimizing input prompts rather than modifying the model's parameters. TPT and DiffTPT adapt the prompt at test time, with DiffTPT also introducing more diverse test sample augmentation using a diffusion model. Historical Prompt Tuning uses prior task knowledge to refine prompts for better task adaptation. (3) Efficient test-time adaptation methods, which do not require backpropagation and rely on a cache of representative samples for adaptation, as demonstrated in \citep{farina2024frustratingly,karmanov2024efficient,zhang2024dual,zhangboostadapter}. The results for the above methods are obtained from the original papers.

\begin{table}[!htbp]
\centering
\begin{minipage}[t]{0.6\linewidth}
\caption{Performance under the NDS scenario.}
  \centering
  \resizebox{\linewidth}{!}{
  \begin{tabular}{lcccc>{\columncolor{gray!20}}c}
    \toprule
    {Method}    & ImageNet   &  ImageNet-A   & ImageNet-R & ImageNet-S  & Average      \\
    \midrule
    Zero-Shot  & 68.34 & \ 49.89 & \ 77.65 & \ 48.24 & \ 61.03              \\
    TPT \cite{shu2022test}   &  68.98  & 54.77        &  77.06   &  47.94   &   62.19 \\
    DiffTPT \cite{feng2023diverse}    &  {70.30}  & 55.68     &  75.00   &  46.80   &   61.95     \\
    TDA\cite{karmanov2024efficient} &  69.51 & {60.11} & {80.24} & {50.54} & {65.10} \\
    ZERO \cite{farina2024frustratingly}& 69.31 & 59.61 & 77.22 & 48.40 & 63.64\\
    \cta  & {70.69} & {61.50}   & {81.21} & {51.84} & {66.31} \\
    \bottomrule
    \label{tab:ood-main}
  \end{tabular}
  }
\end{minipage}%
\hfill
\begin{minipage}[t]{0.37\linewidth}
  \caption{Performance comparison across datasets on PLIP model.}
  \centering
  \resizebox{\linewidth}{!}{
  \begin{tabular}{cccc}
    \toprule
    {Dataset} & {Zero-Shot} & {TDA} & {CTA} \\
    \hline
    {Kather} & 45.60 & 49.35 & 61.92 \\
    PanNuke & 69.49 & 69.70 & 74.68 \\
    WSSS4LUAD & 70.31 & 71.96 & 75.07 \\
    average & 61.80 & 63.67 & 70.56 \\
    \bottomrule
    \label{tab:dataset_comparison_plip}
  \end{tabular}
  }
\end{minipage}
\end{table}

\subsection{Comparison with state-of-the-arts methods}


\textbf{Results under the cross-domain generalization scenario.} We first compare \cta with state-of-the-art methods under the cross-domain generalization scenario across 10 diverse image classification datasets, each from a distinct domain with different classes. Tab.~\ref{tab:fine-grained} presents the experimental results. The proposed method achieved the best performance on most datasets and the top two performance on all datasets. For example, when using the ViT-B/16 backbone network, the average performance was improved by 1.03\%. 

\textbf{Results under the natural distribution shifts scenario.} We then compare \cta with state-of-the-art methods in the context of natural distribution shifts. Tab.~\ref{tab:ood-main} presents the experimental results, revealing the following key observation. Leveraging distribution modeling of the representation of test data, \cta achieves superior performance without requiring gradient backpropagation. For instance, using the CLIP-ViT-B/16 backbone network, \cta outperforms the second-best method by an average of 1.21\%, achieving state-of-the-art results across all datasets.

\textbf{Performance validation on pathological image classification with PLIP.} To validate the performance of the proposed method across other pretrained models and application scenarios, we conducted experiments using the PLIP model in the context of pathological image classification. The results are presented in Tab.~\ref{tab:dataset_comparison_plip}. As shown in the table, the proposed method achieved superior performance, with an average improvement of 6.89\% over TDA.



\begin{table}[!h]
  \centering
  \begin{minipage}{0.42\textwidth}
    \centering
      \caption{Comparisons of our \cta with other methods in terms of efficiency (\textit{Testing Time}) and effectiveness (\textit{Accuracy}).}
    \resizebox{\linewidth}{!}{ 
      \begin{tabular}{lcccc}
        \toprule
        Method & Testing Time & Accuracy & Gain \\ \midrule
        \rowcolor{gray!10}  Zero-Shot & 11.82min & 68.34 & 0 \\
        \midrule
        TPT & 447min & 68.98 & +0.64 \\
        DiffTPT & 1346min & \underline{70.30} & \underline{+1.96} \\
        TDA & \textbf{22min} & 69.51 & +1.17 \\
        \cta (Ours) & \textbf{22min} & \textbf{70.69} & \textbf{+2.35} \\ \bottomrule
        \label{tab:models-time-consuming}
      \end{tabular}
    }
  \end{minipage} \hspace{1cm} 
  \begin{minipage}{0.48\textwidth}
  \caption{Comparisons of our \cta with other methods on the ImageNetV2 dataset, where each class contains only 10 samples.}
    \centering
    \resizebox{\linewidth}{!}{ 
      \begin{tabular}{lccc}
        \toprule
        Method & ViT-B/16 & ResNet-50  \\ \midrule
        \rowcolor{gray!10}  Zero-Shot & 61.88 & 52.91 \\
        \midrule
        TDA & 64.67 & 55.54 \\
        \cta (All test samples) & 64.50 & 55.19 \\ 
        \cta (last 50\% test samples ) & 65.20 & 55.72 \\ 
        \bottomrule
        \label{tab:imagenetv2}
      \end{tabular}
    }
  \end{minipage}
\end{table}

\textbf{Inference time comparison.} To illustrate the efficiency of the proposed method, we conduct evaluation about the inference time using the ViT-B/16 backbone on the ImageNet \citep{deng2009imagenet} dataset. The experimental results are shown in Tab.~\ref{tab:models-time-consuming}. From the table, we can see that the proposed method is faster than the methods that require gradient backpropagation. For example, \cta is 24 times faster than TPT, and 61 times faster than DiffTPT. Therefore, test-time adaptation methods that require gradient backpropagation may not be applicable during deployment due to the performance limitations of the inference device. At the same time, compared with TDA, the speed of the proposed method is comparable, but the performance is higher. 

\textbf{Failure case study.} While our approach highlights the advantages of continuously estimating the distribution of test data and adapting to it, it does not consistently outperform TDA across all datasets, particularly those with a limited number of samples. For instance, as shown in Tab.~\ref{tab:imagenetv2}, on the ImagenetV2 \citep{recht2019imagenet}, which contains only 10 samples per class, the performance of \cta is slightly lower than TDA. This is likely because the limited number of samples per class is insufficient to accurately estimate the data distribution online. However, its performance shows a marked improvement on the latter 50\% of the test samples. This observation suggests that the proposed model has the potential for further enhancement as the number of available test samples increases. 

\begin{table*}[!ht]
  \caption{Performance of \cta and TDA, comparing overall accuracy and the last 50\% of test samples to show continuous adaptability. The results show that the performance of our method is continual improving. However, TDA is different, and its performance has declined on several datasets.}
  \centering
  \resizebox{1\linewidth}{!}{
    \begin{tabular}{l*{10}c>{\columncolor{gray!10}}c}
      \toprule
      Method & Aircraft & Caltech101 & Cars & DTD  & Flower102 & Food101 & Pets & SUN397 & UCF101  \\
      \midrule
    TDA (all test samples) & 23.91 & 94.24 & 67.28 & 47.40 & 71.42 & 86.14 & 88.63 & 67.62 & 70.66 \\
     {TDA (last 50\% test samples)} &  26.57	&93.59 &	66.95&	46.22	&71.75	&86.02&	89.26&	67.86&	72.20  \\    
     \midrule
     \cta (All test samples) &  26.25 & 94.16 & 69.56 & 47.64 & 75.23 & 87.08 & 92.01 & 69.80  & 72.54 \\
   
     {\cta (last 50\% test samples)} & 27.65 & 94.65 & 69.98 & 50.47 & 76.46 & 87.12 & 93.30  & 70.69 & 73.73  \\

      \bottomrule
    \end{tabular}
  } 
  \label{tab:continual_improve}
\end{table*}


\subsection{Ablation studies and further analysis}

\begin{wrapfigure}[16]{r}{0.45\textwidth}
    \centering
\includegraphics[width=\linewidth,clip=false,trim=10 10 10 10]{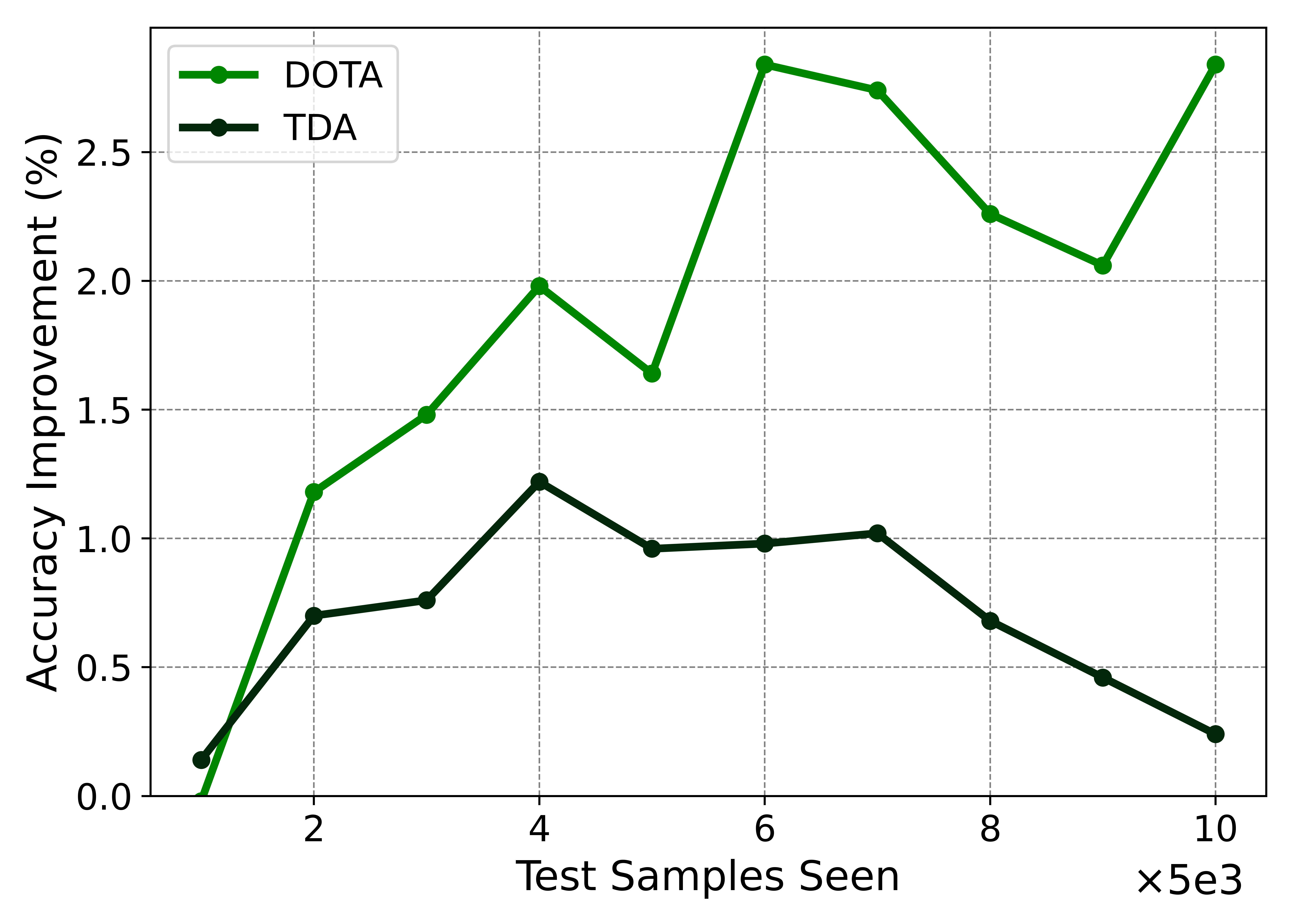} 
\vspace{-2em}
    \caption{Improvement of different methods in model performance as the number of encountered test samples increases.}
    \label{fig:imagenet_improvement}
\end{wrapfigure}

\textbf{Analysis of continuous learning ability and test-time forgetting of TDA.} We conducted this experiment to validate our research motivation and demonstrate experimentally that \cta possesses the capability for continuous learning. During testing on the ImageNet dataset, we recorded the performance of the most recent 5,000 test samples and compared it with the original zero-shot classifier's performance. We analyzed the relationship between the improvement in model performance and the number of test samples processed. The results are illustrated in Fig.~\ref{fig:imagenet_improvement}. From the experimental results, it can be shown that the proposed method progressively enhances model performance as the number of test samples increases. In contrast, TDA shows an initial improvement that subsequently declines, indicating its inability to continuously learn from the test data stream. Further analysis on additional datasets, shown in Tab.~\ref{tab:continual_improve}, highlights the performance on the last 50\% of test samples as well as all samples. The results clearly demonstrate that the performance of the last 50\% of test samples for \cta is significantly higher than the overall performance. This improvement can be attributed to the increasing reliability of the estimated distribution as more test samples are observed. However, TDA exhibits a different pattern, with its performance declining on several datasets.

\setlength{\intextsep}{5pt}



\begin{table*}[!ht]
  \caption{Ablation study to compare the performance of \cta with two variants: (1) rely solely on high-confidence samples and their predictive labels for estimating the distribution; (2) using only the mean, without updating the covariance matrix in the estimation of the Gaussian distribution.}
  \centering
  \resizebox{1\linewidth}{!}{
\begin{tabular}{l*{10}c>{\columncolor{gray!10}}c}
  \toprule
  Method & Aircraft & Caltech101 & Cars & DTD & EuroSAT & Flower102 & Food101 & Pets & SUN397 & UCF101 & Average \\
  \midrule
  \cta  & 26.25 & 94.16 & 69.56 & 47.64 & 62.78 & 75.23 & 87.08 & 92.01 & 69.80  & 72.54 & 69.71  \\
   \midrule						
  \multirow{2}{*}{high-confidence samples only} &  24.63 & 94.12 & 67.11 & 46.34 & 53.28 & 72.07 & 86.47 & 90.68 & 68.24 & 69.28 & 67.22  \\
   &  \textcolor{brown}{-1.62}  &  \textcolor{brown}{-0.04}  &  \textcolor{brown}{-2.45}  &  \textcolor{brown}{-1.30}  &  \textcolor{brown}{-9.50}  &  \textcolor{brown}{-3.16}  &  \textcolor{brown}{-0.61}  &  \textcolor{brown}{-1.33}  &  \textcolor{brown}{-1.56}  &  \textcolor{brown}{-3.26} &  \textcolor{brown}{-2.49}  \\
  \midrule
 \multirow{2}{*}{w/o covariance} &  25.29 & 94.16 & 67.47 & 45.62 & 55.06 & 71.34 & 86.44 & 90.57 & 67.88 & 69.34 & 67.32 \\
   &  \textcolor{brown}{-0.96}  &  \textcolor{black}{0.00}  &  \textcolor{brown}{-2.09}  &  \textcolor{brown}{-2.02}  &  \textcolor{brown}{-7.72}  &  \textcolor{brown}{-3.89}  &  \textcolor{brown}{-0.64}  &  \textcolor{brown}{-1.44}  &  \textcolor{brown}{-1.92}  &  \textcolor{brown}{-3.20} &  \textcolor{brown}{-2.39}  \\

  \bottomrule
\end{tabular}
}
  \label{tab:ablation_cd_mean}
\end{table*}

\textbf{The Importance of distribution estimation using predictive probabilities of all test samples in an EM framework.} We compared the performance of the \cta with a simplified version that only uses high-confidence samples and their corresponding predictive labels for estimating the distribution. The experimental results are shown in Tab.~\ref{tab:ablation_cd_mean}. The experimental results demonstrate that updating the data distribution using the proposed zero-shot predictive probabilities achieves better performance compared to using pseudo-labels from only high-confidence samples. For instance, overall performance sees a decline of approximately 2.49\% when relying solely on high-confidence pseudo-label.

\textbf{The necessity of distribution estimation.}  We compared the performance of the \cta with a simplified version that uses only the mean, excluding the estimation of the Gaussian distribution by removing the updates to the covariance matrices. This experiment aimed to understand the necessity of continual distribution estimation in enhancing model accuracy. The experimental results are shown in Tab.~\ref{tab:ablation_cd_mean}. The fifth row in the table presents the accuracy reductions across different datasets when the covariance matrix is not updated. The results indicate a consistent decrease in accuracy across all datasets, with a particularly notable drop of 7.72\% on the EuroSAT dataset. These findings highlight the importance of continual distribution estimation.

\begin{wraptable}[13]{r}{0.45\textwidth} 
  \centering
\centering
\begin{minipage}{\linewidth}
\centering
\resizebox{1\linewidth}{!}{
\begin{tabular}{c|cccccc}
\toprule
$\sigma^2$ & 0.0001 & 0.001 & 0.002 & 0.004 & 0.008 & 0.02 \\ \midrule
Acc & 70.72 & 70.72 & 70.69 & 70.70  & 70.60  & 70.42  \\
\bottomrule
\end{tabular}
}
\end{minipage}
\begin{minipage}{\linewidth}
\centering
\resizebox{1\linewidth}{!}{
\begin{tabular}{c|cccc}
\toprule
$\eta \backslash \rho$ & 0.005 & 0.01 & 0.02 & 0.03 \\ \midrule
0.2   & 70.69 & 70.66 & 70.59 & 70.54 \\
    0.3   & 70.64 & 70.55 & 70.36 & 70.28 \\
    0.4   & 70.64 & 70.51 & 70.24 & 70.13 \\
    0.5   & 70.64 & 70.48 & 70.15 & 70.00 \\
\bottomrule
\end{tabular}
}
\end{minipage}
\caption{Hyperparameters analysis on the $\sigma^2$ and \((\rho, \eta)\) combinations.}
\label{table:combined}
\end{wraptable}
\textbf{Hyperparameters analysis.} To validate the sensitivity of our model to hyperparameters, we conduct systematic experiments and analyses. First, we evaluate the hyperparameter $\sigma^2$ while keeping other fixed. The results showed minimal impact on accuracy, with performance ranging from 70.42 to 70.72. Next, we test different $\rho$ and $\eta$ combinations, observing stable performance across combinations. For instance, accuracy ranged from 70.69 to 70.00 as $\rho$ and $\eta$ varied. Another hyperparameter $\omega$ we consistently set to 0.0001.  Notably, all hyperparameter combinations show that the proposed method outperforms the original zero-shot classifier, indicating that TTA can usually significantly enhance performance even without a validation set for hyperparameter tuning.

\textbf{The necessity of adaptive fusion of zero-shot and test-time classifier}. We conduct ablation study to show that adaptive fusion of zero-shot and test-time classifier is necessary. The specific experimental results are shown on the Tab.~\ref{table:combined}. It can be observed that as \( \rho \) increases (indicating the diminishing effect of dynamic fusion), the performance of \cta consistently decreases.



\section{Conclusion}

In this paper, we proposed DistributiOnal Test-time Adaptation (\cta), a method that overcomes the limitations of cache-based test-time adaptation by continuously estimating the underlying distribution of test data. Unlike traditional methods, which are constrained by fixed cache sizes and suffer from catastrophic forgetting, \cta dynamically updates class distributions, leading to more efficient and adaptive test-time inference. Our experiments show that \cta significantly reduces forgetting, improves performance, and offers over 20 times faster inference compared to test-time prompt training methods. This approach provides an effective solution for deploying vision-language models in dynamic environments, offering a scalable and computationally efficient alternative to traditional adaptation techniques.

\begin{ack}
This work is supported by the National Natural Science
Foundation of China (624B2100). We thank Prof. Qinghua Hu for his help with the manuscript. Mike Shou does not receive any funding for this work. The authors appreciate the valuable feedback from anonymous reviewers.
\end{ack}
\bibliography{example_paper}

\begin{thebibliography}{10}

\bibitem{anderson1958introduction}
Theodore~Wilbur Anderson, Theodore~Wilbur Anderson, Theodore~Wilbur Anderson, and Theodore~Wilbur Anderson.
\newblock {\em An introduction to multivariate statistical analysis}, volume~2.
\newblock Wiley New York, 1958.

\bibitem{bai2024id}
Yichen Bai, Zongbo Han, Bing Cao, Xiaoheng Jiang, Qinghua Hu, and Changqing Zhang.
\newblock Id-like prompt learning for few-shot out-of-distribution detection.
\newblock In {\em Proceedings of the IEEE/CVF Conference on Computer Vision and Pattern Recognition}, pages 17480--17489, 2024.

\bibitem{bendale2015towards}
Abhijit Bendale and Terrance Boult.
\newblock Towards open world recognition.
\newblock In {\em Proceedings of the IEEE conference on computer vision and pattern recognition}, pages 1893--1902, 2015.

\bibitem{bossard2014food}
Lukas Bossard, Matthieu Guillaumin, and Luc Van~Gool.
\newblock Food-101--mining discriminative components with random forests.
\newblock In {\em Computer vision--ECCV 2014: 13th European conference, zurich, Switzerland, September 6-12, 2014, proceedings, part VI 13}, pages 446--461. Springer, 2014.

\bibitem{boudiaf2022parameter}
Malik Boudiaf, Romain Mueller, Ismail Ben~Ayed, and Luca Bertinetto.
\newblock Parameter-free online test-time adaptation.
\newblock In {\em Proceedings of the IEEE/CVF Conference on Computer Vision and Pattern Recognition}, pages 8344--8353, 2022.

\bibitem{chen2022contrastive}
Dian Chen, Dequan Wang, Trevor Darrell, and Sayna Ebrahimi.
\newblock Contrastive test-time adaptation.
\newblock In {\em Proceedings of the IEEE/CVF Conference on Computer Vision and Pattern Recognition}, pages 295--305, 2022.

\bibitem{cimpoi2014describing}
Mircea Cimpoi, Subhransu Maji, Iasonas Kokkinos, Sammy Mohamed, and Andrea Vedaldi.
\newblock Describing textures in the wild.
\newblock In {\em Proceedings of the IEEE conference on computer vision and pattern recognition}, pages 3606--3613, 2014.

\bibitem{dasgupta2007line}
Sanjoy Dasgupta and Daniel Hsu.
\newblock On-line estimation with the multivariate gaussian distribution.
\newblock In {\em International Conference on Computational Learning Theory}, pages 278--292. Springer, 2007.

\bibitem{de2021continual}
Matthias De~Lange and Tinne Tuytelaars.
\newblock Continual prototype evolution: Learning online from non-stationary data streams.
\newblock In {\em Proceedings of the IEEE/CVF international conference on computer vision}, pages 8250--8259, 2021.

\bibitem{deng2009imagenet}
Jia Deng, Wei Dong, Richard Socher, Li-Jia Li, Kai Li, and Li~Fei-Fei.
\newblock Imagenet: A large-scale hierarchical image database.
\newblock In {\em 2009 IEEE conference on computer vision and pattern recognition}, pages 248--255. Ieee, 2009.

\bibitem{farina2024frustratingly}
Matteo Farina, Gianni Franchi, Giovanni Iacca, Massimiliano Mancini, and Elisa Ricci.
\newblock Frustratingly easy test-time adaptation of vision-language models.
\newblock {\em arXiv preprint arXiv:2405.18330}, 2024.

\bibitem{fei2004learning}
Li~Fei-Fei, Rob Fergus, and Pietro Perona.
\newblock Learning generative visual models from few training examples: An incremental bayesian approach tested on 101 object categories.
\newblock In {\em 2004 conference on computer vision and pattern recognition workshop}, pages 178--178. IEEE, 2004.

\bibitem{feng2023diverse}
Chun-Mei Feng, Kai Yu, Yong Liu, Salman Khan, and Wangmeng Zuo.
\newblock Diverse data augmentation with diffusions for effective test-time prompt tuning.
\newblock In {\em Proceedings of the IEEE/CVF International Conference on Computer Vision}, pages 2704--2714, 2023.

\bibitem{friedman1989regularized}
Jerome~H Friedman.
\newblock Regularized discriminant analysis.
\newblock {\em Journal of the American statistical association}, 84(405):165--175, 1989.

\bibitem{gamper2019pannuke}
Jevgenij Gamper, Navid Alemi~Koohbanani, Ksenija Benet, Ali Khuram, and Nasir Rajpoot.
\newblock Pannuke: an open pan-cancer histology dataset for nuclei instance segmentation and classification.
\newblock In {\em Digital Pathology: 15th European Congress, ECDP 2019, Warwick, UK, April 10--13, 2019, Proceedings 15}, pages 11--19. Springer, 2019.

\bibitem{gao2024clip}
Peng Gao, Shijie Geng, Renrui Zhang, Teli Ma, Rongyao Fang, Yongfeng Zhang, Hongsheng Li, and Yu~Qiao.
\newblock Clip-adapter: Better vision-language models with feature adapters.
\newblock {\em International Journal of Computer Vision}, 132(2):581--595, 2024.

\bibitem{han2022wsss4luad}
Chu Han, Xipeng Pan, Lixu Yan, Huan Lin, Bingbing Li, Su~Yao, Shanshan Lv, Zhenwei Shi, Jinhai Mai, Jiatai Lin, et~al.
\newblock Wsss4luad: Grand challenge on weakly-supervised tissue semantic segmentation for lung adenocarcinoma.
\newblock {\em arXiv preprint arXiv:2204.06455}, 2022.

\bibitem{hastie1996discriminant}
Trevor Hastie and Robert Tibshirani.
\newblock Discriminant analysis by gaussian mixtures.
\newblock {\em Journal of the Royal Statistical Society Series B: Statistical Methodology}, 58(1):155--176, 1996.

\bibitem{helber2019eurosat}
Patrick Helber, Benjamin Bischke, Andreas Dengel, and Damian Borth.
\newblock Eurosat: A novel dataset and deep learning benchmark for land use and land cover classification.
\newblock {\em IEEE Journal of Selected Topics in Applied Earth Observations and Remote Sensing}, 12(7):2217--2226, 2019.

\bibitem{hendrycks2021many}
Dan Hendrycks, Steven Basart, Norman Mu, Saurav Kadavath, Frank Wang, Evan Dorundo, Rahul Desai, Tyler Zhu, Samyak Parajuli, Mike Guo, et~al.
\newblock The many faces of robustness: A critical analysis of out-of-distribution generalization.
\newblock In {\em Proceedings of the IEEE/CVF international conference on computer vision}, pages 8340--8349, 2021.

\bibitem{hendrycks2021natural}
Dan Hendrycks, Kevin Zhao, Steven Basart, Jacob Steinhardt, and Dawn Song.
\newblock Natural adversarial examples.
\newblock In {\em Proceedings of the IEEE/CVF conference on computer vision and pattern recognition}, pages 15262--15271, 2021.

\bibitem{huang2023visual}
Zhi Huang, Federico Bianchi, Mert Yuksekgonul, Thomas~J Montine, and James Zou.
\newblock A visual--language foundation model for pathology image analysis using medical twitter.
\newblock {\em Nature Medicine}, pages 1--10, 2023.

\bibitem{iwasawa2021test}
Yusuke Iwasawa and Yutaka Matsuo.
\newblock Test-time classifier adjustment module for model-agnostic domain generalization.
\newblock {\em Advances in Neural Information Processing Systems}, 34:2427--2440, 2021.

\bibitem{karmanov2024efficient}
Adilbek Karmanov, Dayan Guan, Shijian Lu, Abdulmotaleb El~Saddik, and Eric Xing.
\newblock Efficient test-time adaptation of vision-language models.
\newblock In {\em Proceedings of the IEEE/CVF Conference on Computer Vision and Pattern Recognition}, pages 14162--14171, 2024.

\bibitem{kather2019predicting}
Jakob~Nikolas Kather, Johannes Krisam, Pornpimol Charoentong, Tom Luedde, Esther Herpel, Cleo-Aron Weis, Timo Gaiser, Alexander Marx, Nektarios~A Valous, Dyke Ferber, et~al.
\newblock Predicting survival from colorectal cancer histology slides using deep learning: A retrospective multicenter study.
\newblock {\em PLoS medicine}, 16(1):e1002730, 2019.

\bibitem{khattak2023maple}
Muhammad~Uzair Khattak, Hanoona Rasheed, Muhammad Maaz, Salman Khan, and Fahad~Shahbaz Khan.
\newblock Maple: Multi-modal prompt learning.
\newblock In {\em Proceedings of the IEEE/CVF Conference on Computer Vision and Pattern Recognition}, pages 19113--19122, 2023.

\bibitem{khurana2021sita}
Ansh Khurana, Sujoy Paul, Piyush Rai, Soma Biswas, and Gaurav Aggarwal.
\newblock Sita: Single image test-time adaptation.
\newblock {\em arXiv preprint arXiv:2112.02355}, 2021.

\bibitem{krause20133d}
Jonathan Krause, Michael Stark, Jia Deng, and Li~Fei-Fei.
\newblock 3d object representations for fine-grained categorization.
\newblock In {\em Proceedings of the IEEE international conference on computer vision workshops}, pages 554--561, 2013.

\bibitem{DBLP:conf/icml/LavoieKIAWCB24}
Samuel Lavoie, Polina Kirichenko, Mark Ibrahim, Mido Assran, Andrew~Gordon Wilson, Aaron~C. Courville, and Nicolas Ballas.
\newblock Modeling caption diversity in contrastive vision-language pretraining.
\newblock In {\em Forty-first International Conference on Machine Learning, {ICML} 2024, Vienna, Austria, July 21-27, 2024}. OpenReview.net, 2024.

\bibitem{li2024graphadapter}
Xin Li, Dongze Lian, Zhihe Lu, Jiawang Bai, Zhibo Chen, and Xinchao Wang.
\newblock Graphadapter: Tuning vision-language models with dual knowledge graph.
\newblock {\em Advances in Neural Information Processing Systems}, 36, 2024.

\bibitem{lim2023ttn}
Hyesu Lim, Byeonggeun Kim, Jaegul Choo, and Sungha Choi.
\newblock Ttn: A domain-shift aware batch normalization in test-time adaptation.
\newblock {\em arXiv preprint arXiv:2302.05155}, 2023.

\bibitem{maji2013fine}
Subhransu Maji, Esa Rahtu, Juho Kannala, Matthew Blaschko, and Andrea Vedaldi.
\newblock Fine-grained visual classification of aircraft.
\newblock {\em arXiv preprint arXiv:1306.5151}, 2013.

\bibitem{marsden2024universal}
Robert~A Marsden, Mario D{\"o}bler, and Bin Yang.
\newblock Universal test-time adaptation through weight ensembling, diversity weighting, and prior correction.
\newblock In {\em Proceedings of the IEEE/CVF Winter Conference on Applications of Computer Vision}, pages 2555--2565, 2024.

\bibitem{nado2020evaluating}
Zachary Nado, Shreyas Padhy, D~Sculley, Alexander D'Amour, Balaji Lakshminarayanan, and Jasper Snoek.
\newblock Evaluating prediction-time batch normalization for robustness under covariate shift.
\newblock {\em arXiv preprint arXiv:2006.10963}, 2020.

\bibitem{nilsback2008automated}
Maria-Elena Nilsback and Andrew Zisserman.
\newblock Automated flower classification over a large number of classes.
\newblock In {\em 2008 Sixth Indian conference on computer vision, graphics \& image processing}, pages 722--729. IEEE, 2008.

\bibitem{parkhi2012cats}
Omkar~M Parkhi, Andrea Vedaldi, Andrew Zisserman, and CV~Jawahar.
\newblock Cats and dogs.
\newblock In {\em 2012 IEEE conference on computer vision and pattern recognition}, pages 3498--3505. IEEE, 2012.

\bibitem{pratt2023does}
Sarah Pratt, Ian Covert, Rosanne Liu, and Ali Farhadi.
\newblock What does a platypus look like? generating customized prompts for zero-shot image classification.
\newblock In {\em Proceedings of the IEEE/CVF International Conference on Computer Vision}, pages 15691--15701, 2023.

\bibitem{radford2021learning}
Alec Radford, Jong~Wook Kim, Chris Hallacy, Aditya Ramesh, Gabriel Goh, Sandhini Agarwal, Girish Sastry, Amanda Askell, Pamela Mishkin, Jack Clark, et~al.
\newblock Learning transferable visual models from natural language supervision.
\newblock In {\em International conference on machine learning}, pages 8748--8763. PMLR, 2021.

\bibitem{recht2019imagenet}
Benjamin Recht, Rebecca Roelofs, Ludwig Schmidt, and Vaishaal Shankar.
\newblock Do imagenet classifiers generalize to imagenet?
\newblock In {\em International conference on machine learning}, pages 5389--5400. PMLR, 2019.

\bibitem{shu2022test}
Manli Shu, Weili Nie, De-An Huang, Zhiding Yu, Tom Goldstein, Anima Anandkumar, and Chaowei Xiao.
\newblock Test-time prompt tuning for zero-shot generalization in vision-language models.
\newblock {\em Advances in Neural Information Processing Systems}, 35:14274--14289, 2022.

\bibitem{snell2017prototypical}
Jake Snell, Kevin Swersky, and Richard Zemel.
\newblock Prototypical networks for few-shot learning.
\newblock {\em Advances in neural information processing systems}, 30, 2017.

\bibitem{soomro2012dataset}
Khurram Soomro, Amir~Roshan Zamir, and Mubarak Shah.
\newblock A dataset of 101 human action classes from videos in the wild.
\newblock {\em Center for Research in Computer Vision}, 2(11):1--7, 2012.

\bibitem{wang2020tent}
Dequan Wang, Evan Shelhamer, Shaoteng Liu, Bruno Olshausen, and Trevor Darrell.
\newblock Tent: Fully test-time adaptation by entropy minimization.
\newblock {\em arXiv preprint arXiv:2006.10726}, 2020.

\bibitem{DBLP:conf/iclr/WangSLOD21}
Dequan Wang, Evan Shelhamer, Shaoteng Liu, Bruno~A. Olshausen, and Trevor Darrell.
\newblock Tent: Fully test-time adaptation by entropy minimization.
\newblock In {\em 9th International Conference on Learning Representations, {ICLR} 2021, Virtual Event, Austria, May 3-7, 2021}. OpenReview.net, 2021.

\bibitem{wang2019learning}
Haohan Wang, Songwei Ge, Zachary Lipton, and Eric~P Xing.
\newblock Learning robust global representations by penalizing local predictive power.
\newblock {\em Advances in Neural Information Processing Systems}, 32, 2019.

\bibitem{DBLP:conf/iclr/WangLS0WT24}
Zhengbo Wang, Jian Liang, Lijun Sheng, Ran He, Zilei Wang, and Tieniu Tan.
\newblock A hard-to-beat baseline for training-free clip-based adaptation.
\newblock In {\em The Twelfth International Conference on Learning Representations, {ICLR} 2024, Vienna, Austria, May 7-11, 2024}. OpenReview.net, 2024.

\bibitem{xiao2010sun}
Jianxiong Xiao, James Hays, Krista~A Ehinger, Aude Oliva, and Antonio Torralba.
\newblock Sun database: Large-scale scene recognition from abbey to zoo.
\newblock In {\em 2010 IEEE computer society conference on computer vision and pattern recognition}, pages 3485--3492. IEEE, 2010.

\bibitem{yu2023task}
Tao Yu, Zhihe Lu, Xin Jin, Zhibo Chen, and Xinchao Wang.
\newblock Task residual for tuning vision-language models.
\newblock In {\em Proceedings of the IEEE/CVF Conference on Computer Vision and Pattern Recognition}, pages 10899--10909, 2023.

\bibitem{yuan2023robust}
Longhui Yuan, Binhui Xie, and Shuang Li.
\newblock Robust test-time adaptation in dynamic scenarios.
\newblock In {\em Proceedings of the IEEE/CVF Conference on Computer Vision and Pattern Recognition}, pages 15922--15932, 2023.

\bibitem{zhai2023sigmoid}
Xiaohua Zhai, Basil Mustafa, Alexander Kolesnikov, and Lucas Beyer.
\newblock Sigmoid loss for language image pre-training.
\newblock In {\em Proceedings of the IEEE/CVF International Conference on Computer Vision}, pages 11975--11986, 2023.

\bibitem{zhang2024historical}
Jingyi Zhang, Jiaxing Huang, Xiaoqin Zhang, Ling Shao, and Shijian Lu.
\newblock Historical test-time prompt tuning for vision foundation models.
\newblock {\em arXiv preprint arXiv:2410.20346}, 2024.

\bibitem{zhang2022memo}
Marvin Zhang, Sergey Levine, and Chelsea Finn.
\newblock Memo: Test time robustness via adaptation and augmentation.
\newblock {\em Advances in neural information processing systems}, 35:38629--38642, 2022.

\bibitem{zhang2022tip}
Renrui Zhang, Wei Zhang, Rongyao Fang, Peng Gao, Kunchang Li, Jifeng Dai, Yu~Qiao, and Hongsheng Li.
\newblock Tip-adapter: Training-free adaption of clip for few-shot classification.
\newblock In {\em European conference on computer vision}, pages 493--510. Springer, 2022.

\bibitem{zhang2024boostadapter}
Taolin Zhang, Jinpeng Wang, Hang Guo, Tao Dai, Bin Chen, and Shu-Tao Xia.
\newblock Boostadapter: Improving vision-language test-time adaptation via regional bootstrapping.
\newblock {\em arXiv preprint arXiv:2410.15430}, 2024.

\bibitem{zhangboostadapter}
Taolin Zhang, Jinpeng Wang, Hang Guo, Tao Dai, Bin Chen, and Shu-Tao Xia.
\newblock Boostadapter: Improving vision-language test-time adaptation via regional bootstrapping.
\newblock In {\em The Thirty-eighth Annual Conference on Neural Information Processing Systems}, 2024.

\bibitem{zhang2024dual}
Yabin Zhang, Wenjie Zhu, Hui Tang, Zhiyuan Ma, Kaiyang Zhou, and Lei Zhang.
\newblock Dual memory networks: A versatile adaptation approach for vision-language models.
\newblock In {\em Proceedings of the IEEE/CVF conference on computer vision and pattern recognition}, pages 28718--28728, 2024.

\bibitem{zhou2022cocoop}
Kaiyang Zhou, Jingkang Yang, Chen~Change Loy, and Ziwei Liu.
\newblock Conditional prompt learning for vision-language models.
\newblock In {\em IEEE/CVF Conference on Computer Vision and Pattern Recognition (CVPR)}, 2022.

\bibitem{zhou2022coop}
Kaiyang Zhou, Jingkang Yang, Chen~Change Loy, and Ziwei Liu.
\newblock Learning to prompt for vision-language models.
\newblock {\em International Journal of Computer Vision (IJCV)}, 2022.

\end{thebibliography}
\bibliographystyle{plain}

\appendix
\newpage
\section{Appendix}
\subsection{Proof of Proposition~\ref{tta:lemma1}}
\label{proof:tta:lemma1}
The EM algorithm is an iterative method used for finding maximum likelihood estimates of parameters in statistical models with latent (unobserved) variables. It consists of two main steps:

(1) Expectation Step (E-step): Compute the expected value of the log-likelihood function, with respect to the current estimate of the distribution of the latent variables.

(2) Maximization Step (M-step): Maximize this expected log-likelihood to update the parameter estimates.

In the context of Gaussian Discriminant Analysis (GDA) for classification, the latent variables correspond to the true class labels of the data points, which are unobserved during testing.

\begin{assumption}[Class-Conditional Distributions]
For each class \( k \in \{1, 2, \dots, K\} \), the data embedding \( \boldsymbol{x} \) follows a Gaussian distribution:
  \[
  P(\boldsymbol{x} \mid y = k) = \mathcal{N}(\boldsymbol{\mu}_k, \boldsymbol{\Sigma}_k)
  \]
  where \( \boldsymbol{\mu}_k \) and \( \boldsymbol{\Sigma}_k \) are the mean vector and covariance matrix of class \( k \), respectively.
\end{assumption}

\begin{assumption}[Prior Probabilities]
The prior probability for each class is uniform:
  \(
  P(y = k) = \frac{1}{K}.
  \)
\end{assumption}
The objective is how to estimate the parameters \( \{\boldsymbol{\mu}_k, \boldsymbol{\Sigma}_k\}_{k=1}^K \) using the EM algorithm, leveraging the zero-shot predictive probabilities \( P^{\texttt{zs}}_k(y = k \mid \boldsymbol{x}_n) \) as part of the expectation step.

\paragraph{E-Step: Compute Expected Log-Likelihood.} In the E-step, we compute the expectation of the log-likelihood with respect to the posterior distribution of the latent variables given the current parameter estimates. Given the zero-shot predictive probabilities \( P^{\texttt{zs}}_k(y = k \mid \boldsymbol{x}_n) \), we treat them as the responsibilities (i.e., the posterior probabilities) in the E-step:
\[
\pi_{nk} = P^{\texttt{zs}}_k(y = k \mid \boldsymbol{x}_n)
\]
The expected log-likelihood \( Q \) is then:
\[
Q(\{\boldsymbol{\mu}_k, \boldsymbol{\Sigma}_k\} \mid \text{current estimates}) = \sum_{n=1}^N \sum_{k=1}^K \pi_{nk} \left[ \log \mathcal{N}(\boldsymbol{x}_n \mid \boldsymbol{\mu}_k, \boldsymbol{\Sigma}_k) + \log \frac{1}{K} \right]
\]

\paragraph{M-Step: Maximize the Expected Log-Likelihood.} 
In the M-step, we maximize \( Q \) with respect to \( \{\boldsymbol{\mu}_k, \boldsymbol{\Sigma}_k\}_{k=1}^K \).
\textbf{Maximizing with Respect to \( \boldsymbol{\mu}_k \).}
To find the optimal \( \boldsymbol{\mu}_k \), take the derivative of \( Q \) with respect to \( \boldsymbol{\mu}_k \) and set it to zero. First, expand the Gaussian log-probability:
\[
\log \mathcal{N}(\boldsymbol{x}_n \mid \boldsymbol{\mu}_k, \boldsymbol{\Sigma}_k) = -\frac{1}{2} (\boldsymbol{x}_n - \boldsymbol{\mu}_k)^\top \boldsymbol{\Sigma}_k^{-1} (\boldsymbol{x}_n - \boldsymbol{\mu}_k) - \frac{1}{2} \log |\boldsymbol{\Sigma}_k| - \frac{d}{2} \log (2\pi)
\]
where \( d \) is the dimensionality of \( \boldsymbol{x}_n \). Focusing on terms involving \( \boldsymbol{\mu}_k \):
\[
Q_{\mu} = -\frac{1}{2} \sum_{n=1}^N \pi_{nk} (\boldsymbol{x}_n - \boldsymbol{\mu}_k)^\top \boldsymbol{\Sigma}_k^{-1} (\boldsymbol{x}_n - \boldsymbol{\mu}_k)
\]
Take the derivative with respect to \( \boldsymbol{\mu}_k \) and set to zero:
\[
\frac{\partial Q_{\mu}}{\partial \boldsymbol{\mu}_k} = \sum_{n=1}^N \pi_{nk} \boldsymbol{\Sigma}_k^{-1} (\boldsymbol{x}_n - \boldsymbol{\mu}_k) = 0
\]
Solving for \( \boldsymbol{\mu}_k \):
\[
\sum_{n=1}^N \pi_{nk} \boldsymbol{x}_n = \left( \sum_{n=1}^N \pi_{nk} \right) \boldsymbol{\mu}_k
\]
\[
\boldsymbol{\mu}_k = \frac{\sum_{n=1}^N \pi_{nk} \boldsymbol{x}_n}{\sum_{n=1}^N \pi_{nk}}
\]
This corresponds directly to the provided update formula for \( \hat{\boldsymbol{\mu}}_k \):
\[
\hat{\boldsymbol{\mu}}_k = \frac{\sum_{n=1}^N P^{\texttt{zs}}_{k}(y = k \mid \boldsymbol{x}_n) \boldsymbol{x}_n}{\sum_{n=1}^N P^{\texttt{zs}}_{k}(y = k \mid \boldsymbol{x}_n)}
\]

\textbf{Maximizing with Respect to \( \boldsymbol{\Sigma}_k \).} Similarly, to find the optimal \( \boldsymbol{\Sigma}_k \), consider the terms in \( Q \) involving \( \boldsymbol{\Sigma}_k \):
\[
Q_{\Sigma} = -\frac{1}{2} \sum_{n=1}^N \pi_{nk} \left[ (\boldsymbol{x}_n - \boldsymbol{\mu}_k)^\top \boldsymbol{\Sigma}_k^{-1} (\boldsymbol{x}_n - \boldsymbol{\mu}_k) + \log |\boldsymbol{\Sigma}_k| \right]
\]
Take the derivative with respect to \( \boldsymbol{\Sigma}_k^{-1} \) (using matrix derivative identities) and set to zero:
\[
\frac{\partial Q_{\Sigma}}{\partial \boldsymbol{\Sigma}_k^{-1}} = \frac{1}{2} \sum_{n=1}^N \pi_{nk} \left[ (\boldsymbol{x}_n - \boldsymbol{\mu}_k)(\boldsymbol{x}_n - \boldsymbol{\mu}_k)^\top - \boldsymbol{\Sigma}_k \right] = 0
\]
Solving for \( \boldsymbol{\Sigma}_k \):
\[
\boldsymbol{\Sigma}_k = \frac{\sum_{n=1}^N \pi_{nk} (\boldsymbol{x}_n - \boldsymbol{\mu}_k)(\boldsymbol{x}_n - \boldsymbol{\mu}_k)^\top}{\sum_{n=1}^N \pi_{nk}}
\]
This aligns with the provided update formula for \( \hat{\boldsymbol{\Sigma}}_k \):
\[
\hat{\boldsymbol{\Sigma}}_k = \frac{\sum_{n=1}^N P^{\texttt{zs}}_{k}(y = k \mid \boldsymbol{x}_n) (\boldsymbol{x}_n - \hat{\boldsymbol{\mu}}_k)(\boldsymbol{x}_n - \hat{\boldsymbol{\mu}}_k)^\top}{\sum_{n=1}^N P^{\texttt{zs}}_{k}(y = k \mid \boldsymbol{x}_n)}
\]

Finally, the provided update equation is:
\[
\hat{\boldsymbol{\mu}}_k = \frac{\sum_{n=1}^N P^{\texttt{zs}}_{k}(y = k \mid \boldsymbol{x}_n) \boldsymbol{x}_n}{\sum_{n=1}^N P^{\texttt{zs}}_{k}(y = k \mid \boldsymbol{x}_n)}, \quad
\hat{\boldsymbol{\Sigma}}_k = \frac{\sum_{n=1}^N P^{\texttt{zs}}_{k}(y = k \mid \boldsymbol{x}_n) (\boldsymbol{x}_n - \hat{\boldsymbol{\mu}}_k)(\boldsymbol{x}_n - \hat{\boldsymbol{\mu}}_k)^\top}{\sum_{n=1}^N P^{\texttt{zs}}_{k}(y = k \mid \boldsymbol{x}_n)}
\]
are precisely the M-step updates obtained by maximizing the expected log-likelihood in the EM algorithm, where the E-step uses the zero-shot predictive probabilities as responsibilities. This demonstrates that the parameter estimation process described is equivalent to performing a single EM iteration. This single EM iteration leverages the zero-shot predictions to adjust the Gaussian parameters, effectively ``reweighting" the contributions of different samples based on their inferred class probabilities.

\subsection{\textcolor{black}{Results across multiple data ordering.}}
\label{sec:dataodering}
\textcolor{black}{We conducted experiments on five test data in different ordering on multiple datasets. The experimental results are shown in the Tab.~\ref{tab:results_multiorder}. From the experimental results, we can see that the order of test data has little effect on the prediction performance.}
\begin{table}[h!]
    \caption{Experimental results across multiple datasets and test data orderings.}
    \centering
    \textcolor{black}{
    \resizebox{0.8\linewidth}{!}{
    \begin{tabular}{@{}llcccccc@{}}
        \toprule
        Dataset & Method\textbackslash Data ordering & 1 & 2 & 3 & 4 & 5 & Average \\
        \midrule
        ImageNet & \cta & 70.54 & 70.66 & 70.72 & 70.67 & 70.63 & 70.64 \\
        eurosat & \cta & 62.74 & 62.56 & 63.01 & 62.46 & 63.28 & 62.81  \\
        OxfordPets & \cta & 92.04 & 91.93 & 92.29 & 92.04 & 92.12 & 92.08 \\
        \bottomrule
    \end{tabular}
    }
    \label{tab:results_multiorder}
    }
\end{table}
\subsection{Performance on non-i.i.d. data streams}
\textcolor{black}{
We conducted additional experiments to evaluate the model's performance under non-i.i.d. data distribution during testing, using the ImageNet dataset as a benchmark. By employing a Dirichlet distribution, we simulated varying degrees of non-i.i.d. data streams, adjusting the concentration parameter and dividing the dataset into 5 and 10 time slices for analysis. The details of the experiments are shown as follows:}

\textcolor{black}{
\textbf{Time Slices}: We divided the ImageNet dataset into 5 and 10 time slices, where each slice contains varying numbers of samples and class distributions.}

\textcolor{black}{
\textbf{Concentration Parameter ($[\alpha]_K$)}: The concentration parameter of the Dirichlet distribution controls the uniformity of class distributions across slices. Smaller $\alpha$ values (e.g., 0.1) create highly uneven distributions, while larger values (e.g., 0.5 and 1) result in more uniform distributions.}

\textcolor{black}{
\textbf{Evaluation Setting}: Since the sizes of sub-datasets for each time slice are unequal, the final average accuracy is a weighted average based on the number of samples in each slice.
The experimental results are summarized below.}

\begin{table}[htbp]
\centering
\caption{\textcolor{black}{
Performance on non-i.i.d. data streams (5 slices).}}
\textcolor{black}{
\resizebox{0.7\linewidth}{!}{
\begin{tabular}{cccccccc}
\toprule
 $\alpha$ & Slice 1 & Slice 2 & Slice 3 & Slice 4 & Slice 5 & Average \\ \midrule
 0.1      & 68.18 & 70.06 & 71.6  & 71.09 & 70.91 & 70.41    \\ 
 0.5      & 69.55 & 70.78 & 69.43 & 71.95 & 70.92 & 70.58   \\ 
 1        & 69.41 & 71.64 & 71.05 & 70.01 & 71.81 & 70.83   \\ 
\bottomrule
\end{tabular}}}
\label{tab:5_slices}
\end{table}

\begin{table}[htbp]
\caption{\textcolor{black}{
Performance on non-i.i.d. data streams (10 slices).}}
\centering
    \textcolor{black}{
    \resizebox{0.9\linewidth}{!}{
\begin{tabular}{cccccccccccc}
\toprule
 $\alpha$ & Slice 1 & Slice 2 & Slice 3 & Slice 4 & Slice 5 & Slice 6 & Slice 7 & Slice 8 & Slice 9 & Slice 10 & Average \\ \midrule
 0.1      & 69.99 & 69.31 & 67.34 & 70.23 & 71.49 & 68.8  & 73.13 & 70.35 & 69.89 & 72.39 & 70.39 \\
 0.5      & 68.74 & 69.32 & 72.17 & 71.27 & 69.97 & 69.8  & 70.3  & 70.78 & 72.48 & 70.78 & 70.61 \\
 1        & 67.33 & 70.29 & 70.92 & 68.43 & 69.85 & 71.14 & 72.03 & 72.52 & 71.29 & 71.66 & 70.68 \\
 \bottomrule
\end{tabular}}}
\label{tab:10_slices}
\end{table}

\textcolor{black}{From the experimental results, we can see that the model shows strong robustness to non-i.i.d. data streams, with only minimal accuracy decline under small $\alpha$ (e.g., $\alpha=0.1$).}

\subsection{\textcolor{black}{More details and explanation about the $f_k(x)$ in Eq.~\ref{eq:testprob_qda}.}} 
\label{sec:fkxdetail}
\textcolor{black}{The function $f_k(\boldsymbol{x})$, often referred to as the \textit{discriminant function}, measures how well a data point $\boldsymbol{x}$ fits the distribution of class $k$. It is derived from Gaussian Discriminant Analysis and consists of two main components. The first component is the Mahalanobis distance, $-\frac{1}{2} (\boldsymbol{x} - \mu_k)^T \Sigma_k^{-1} (\boldsymbol{x} - \mu_k)$, which calculates the squared distance between $\boldsymbol{x}$ and the class mean $\mu_k$, scaled by the inverse of the covariance matrix $\Sigma_k$. This term captures the similarity of $\boldsymbol{x}$ to the center of the class, considering feature correlations. The second component is the normalization term, $-\frac{1}{2} \log |\Sigma_k|$, which accounts for the determinant of the covariance matrix $\Sigma_k$ and reflects the spread (or volume) of the Gaussian distribution for class $k$. This ensures that classes with larger variances are normalized appropriately. Intuitively, a larger value of $f_k(\boldsymbol{x})$ indicates a higher likelihood that $\boldsymbol{x}$ belongs to class $k$. In classification, $f_k(\boldsymbol{x})$ is used within the softmax function to compute the posterior probability $P(y = k \mid \boldsymbol{x})$, which determines the most likely class for $\boldsymbol{x}$: 
\[P(y = k \mid \boldsymbol{x}) = \frac{\exp(f_k(\boldsymbol{x}))}{\sum_{k=1}^K \exp(f_k(\boldsymbol{x}))}\].}

\begin{table}[h]
\caption{Datasets details.}
    \centering
    \renewcommand\arraystretch{1.0}{
    \begin{tabular}{l c c c l}
    \toprule
    Dataset & Classes & Validation Size & Test Size & \makecell[c]{Task} \\
    \midrule
    ImageNet & 1,000 & N/A   & 50,000 & Classification\\

    ImageNet-V2 & 1,000 & N/A   & 10,000 & Generalization \\

    ImageNet-S & 1,000 & N/A   & 50,000 & Generalization\\
    ImageNet-A & 200   & N/A   & 7,500 &  Generalization \\
    ImageNet-R & 200   & N/A   & 30,000 & Generalization\\
    \midrule
  Aircraft & 100   & 3,333  & 3,333 & Aircraft recognition\\
  Caltech101 & 100   & 1,649  & 2,465 & Object recognition \\
  Cars  & 196   & 1,635  & 8,041 & Car recognition  \\
  DTD   & 47    & 1,128  & 1,692 &  Texture classification \\
  EuroSAT & 10    & 5,400  & 8,100 & Remote sensing classification \\
  Flowers102 & 102   & 1,633  & 2,463 & Flower recognition  \\
  Food101 & 101   & 20,200 & 30,300 & Food classification \\
  Pets  & 37    & 736   & 3,669 & Pet classification\\
  SUN397 & 397   & 3,970  & 19,850 & Scene recognition  \\
 UCF101 & 101   & 1,898  & 3,783 & Action recognition \\
     \midrule
  Kather  & 2   & 10,718  & 32,154 & Colon classification\\
  PanNuke  & 9   & 623  & 1,888 & Tissue classification \\
  WSSS4LUAD   & 2   & 1,009  & 3,028 & Tissue classification  \\
\bottomrule
\end{tabular}}
\label{tab:datasets}
\end{table}

\subsection{Implementation details.}
All the models in our experiments are built upon the pre-trained CLIP model \citep{radford2021learning} that consists of an image encoder and a text encoder. Test-time adaptation is set for single-image scenarios, using a batch size of 1. For  natural distribution shifts scenario, we tune our hyperparameters using the validation set. For the cross-domain generalization scenario, we perform hyperparameter search using the corresponding validation sets. We adjust $\sigma^2$ within [0.001, 0.002, 0.004], then search for the best $\eta$ across [0.2, 0.3, 0.4, 0.5] and $\rho$ across [0.005, 0.01, 0.02, 0.03], with the shrinkage parameter $\epsilon$ set to 0.0001.  We use top-1 accuracy (\%) as our evaluation metric. All experiments are conducted using a single NVIDIA RTX 4090 GPU and a 12-core Intel Xeon Platinum 8352V CPU.

\newpage
\end{document}